\documentclass[lettersize,journal]{IEEEtran}
\usepackage{graphicx} 
\usepackage{booktabs}

\usepackage[pagebackref,breaklinks,colorlinks,citecolor=cvprblue]{hyperref}

\usepackage{makecell}
\usepackage{colortbl}
\usepackage{color}
\definecolor{linkcolor}{RGB}{255,0,0}
\definecolor{urlcolor}{RGB}{255,105,180}
\definecolor{citecolor}{RGB}{66,168,235}
\usepackage{rotating} 
\usepackage{pifont}
\usepackage{amssymb}
\usepackage{multirow}
\usepackage{algorithm}
\usepackage{multicol}
\usepackage{lineno}
\hypersetup{colorlinks=true,linkcolor=linkcolor,urlcolor=urlcolor,citecolor=blue}
\definecolor{tab_others}{RGB}{235, 235, 235}
\definecolor{tab_ours}{RGB}{225, 235, 246}
\usepackage{amsmath}

\title{Learning Multi-view Multi-class Anomaly Detection}
\author{Qianzi Yu, Yang Cao, Yu Kang\\USTC}

\begin{document}

\maketitle
\begin{abstract}
The latest trend in anomaly detection is to train a unified model instead of training a separate model for each category. However, existing multi-class anomaly detection (MCAD) models perform poorly in multi-view scenarios because they often fail to effectively model the relationships and complementary information among different views. In this paper, we introduce a Multi-View Multi-Class Anomaly Detection model (MVMCAD), which integrates information from multiple views to accurately identify anomalies. Specifically, we propose a semi-frozen encoder, where a pre-encoder prior enhancement mechanism is added before the frozen encoder, enabling stable cross-view feature modeling and efficient adaptation for improved anomaly detection. Furthermore, we propose an Anomaly Amplification Module (AAM) that models global token interactions and suppresses normal regions to enhance anomaly signals, leading to improved detection performance in multi-view settings. Finally, we propose a Cross-Feature Loss that aligns shallow encoder features with deep decoder features and vice versa, enhancing the model's sensitivity to anomalies at different semantic levels under multi-view scenarios. Extensive experiments on the Real-IAD dataset for multi-view multi-class anomaly detection validate the effectiveness of our approach, achieving state-of-the-art performance of \textbf{91.0}/\textbf{88.6}/\textbf{82.1} and \textbf{99.1/43.9/48.2/95.2} for image-level and the pixel-level, respectively. 
\end{abstract}

\begin{IEEEkeywords}
Unsupervised Anomaly Detection, Multi-view Learning, Multi-class Anomaly Detection, Encoder-decoder Framework
\end{IEEEkeywords}

\section{Introduction}
\begin{figure}[t]
\centering
\includegraphics[width=\linewidth]{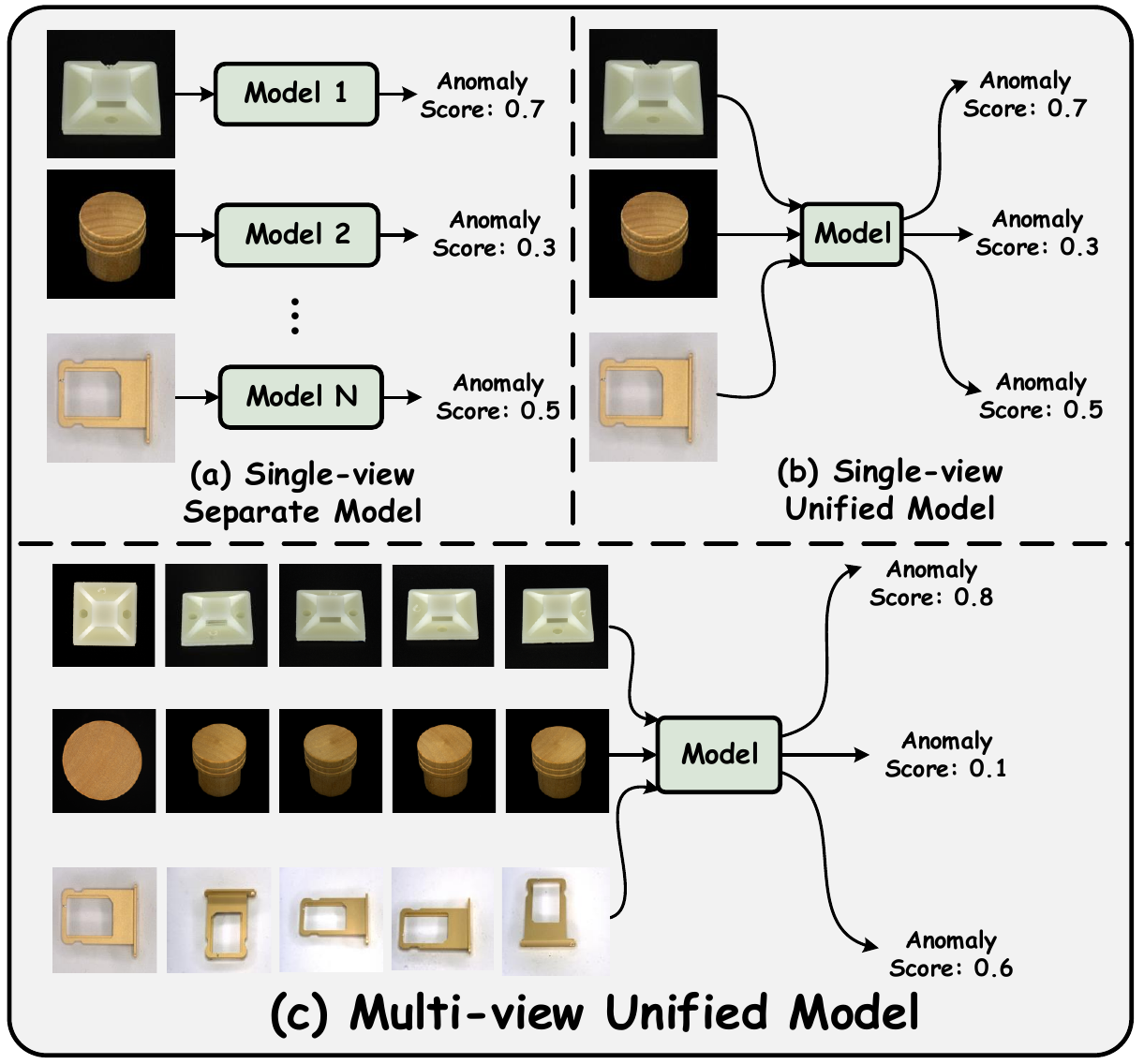}
\caption{\textbf{Task settings.} (a) Task setting of Class-separate, Single-view Unsupervised Anomaly Detection. (b) Task setting of Class-separate, Multi-view Unsupervised Anomaly Detection. (c) Task setting of Multi-class, Multi-view Unsupervised Anomaly Detection.}
\label{first}
\end{figure}

\IEEEPARstart{I}{ndustrial} anomaly detection plays an important role in industrial production, with the aim of identifying and locating anomalies to filter out defective products and adjust the production process. Due to the scarcity of anomaly data, researchers often use unsupervised methods \cite{deng2022anomaly,you2022unified,liu2023simplenet,zhang2023destseg,he2024diffusion,hemambaad,guo2024dinomaly} to train anomaly detection models. They train the model using only normal samples, and examples that deviate from normal samples are considered anomalies. The current trend in industrial image anomaly detection is to train a unified model \cite{you2022unified} rather than training individual models \cite{liu2023simplenet,deng2022anomaly,deng2024prioritized} for each class separately to improve inference speed and simplify model deployment. Moreover, with the introduction of the Real-IAD dataset \cite{wang2024real}, multi-view industrial anomaly detection has started to gain attention. In response to this trend, MVAD \cite{he2024learning} proposes a pioneering framework for multi-view anomaly detection, leveraging a multi-view attention selector to adaptively model cross-view semantic relevance. However, MVAD lacks the ability to handle multi-class scenarios. Therefore, we propose a Multi-View Multi-Class Anomaly Detection (MVMCAD) method, as illustrated in Figure \ref{first}, which unifies multi-class, multi-view anomaly detection within a single framework.

\begin{figure*}[t]
\centering
\includegraphics[width=\linewidth]{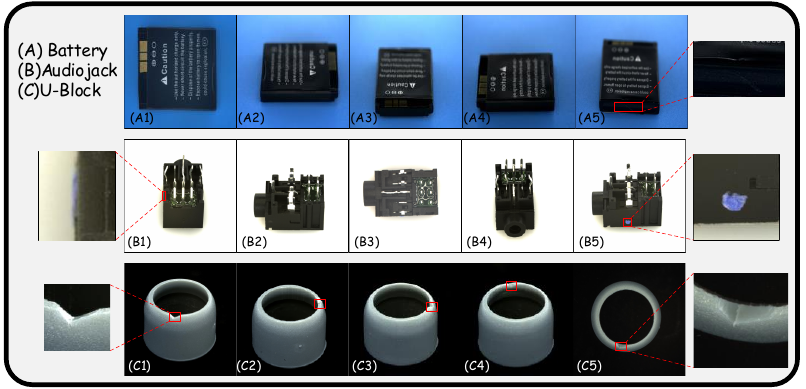}
\caption{\textbf{Challenge of Multi-View Scenarios.} (A) In the phone battery category, the scratch is exclusively visible in (A5) and cannot be observed in views (A1-A4). (B) In the Audiojack category, the contamination in is difficult to detect in view B1, yet it becomes clearly visible in view (B5), highlighting the necessity for the model to exploit inter-view correlations to accurately localize the anomaly in (B1). (C) The third row illustrates that edge-located anomalies in the u-block category tend to be challenging to identify.}
\label{moti}
\end{figure*}

Unlike single-view image anomaly detection, multi-view images provide multiple perspectives of an object. Some existing Multi-Class Anomaly Detection (MCAD) methods perform poorly when handling multi-view images, mainly due to several reasons: (1) Inconsistency of viewpoint data. Due to the correlations between different views, some anomalies may appear in one view but not be visible in others, increasing the difficulty of the task. (2) Insufficient utilization of the correlation between views. Images from different views may have interdependent or shared features, while traditional single-view detection methods cannot fully utilize this cross-view information. (3) Anomalies at the edges of objects are difficult to detect. Anomalies at the edges of objects may have alignment issues between different views. The alignment of cross-view features may become distorted, which in turn decreases the accuracy of anomaly detection. We provide several examples in Figure \ref{moti} to illustrate these challenges more intuitively.

To solve these problems, we introduce an encoder-decoder framework that contains a semi-frozen encoder and an anomaly amplification module. Specifically, the semi-frozen encoder adopts a frozen backbone combined with a fine-tuned pre-encoder prior enhancement mechanism, providing cross-view shared structural adaptation. This trainable mechanism adjusts the low-level feature responses based on image-specific statistics, allowing the encoder to recalibrate the input feature distribution across different views, which enables the model to quickly adapt to the data distributions of different views without sacrificing its original visual representation capability. Moreover, the anomaly amplification module establishes contextual semantics and global token-level associations, while focuses on suppressing tokens that are highly similar to those likely representing normal regions by computing normalized similarity and inversely weighting feature responses to emphasize anomaly or deviant features. In this way, AAM suppresses dominant normal patterns while amplifying semantically deviant or rare tokens, which often correspond to anomalies, which enables the model to uncover anomaly information that may be prominent in one view but easily overlooked in others. Finally, we design a Cross-Feature Loss by connecting shallow features from the encoder to deeper layers in the decoder, and conversely deep features from the encoder to shallow decoder layers. This cross-level feature fusion is motivated by the observation that anomalies in multi-view images may appear at different semantic levels, some anomalies manifest in local textures (low-level), while others only emerge in global structures (high-level). This design enables the decoder to locate anomalies in shallow layers and enhance them in deeper layers, thereby improving the model’s capacity for accurate and robust anomaly detection.

Our main contributions are summarized as follows.
\begin{itemize}
\item [1)] We propose an encoder-decoder framework to train an unified model to solve the multi-view anomaly detection challenge.

\item [2)] A semi-frozen encoder, an anomaly amplification module, and a cross-feature loss are proposed to enable stable cross-view representation learning, enhance the saliency of anomalous regions, and align semantic discrepancies between feature hierarchies for improved multi-view anomaly detection.

\item [3)] Extensive experiments on Real-IAD datasets demonstrate the superiority of our proposed method over SOTA. The metrics indicate that our method performs well on anomaly recognition tasks in both image-level detection and pixel-level localization.
\end{itemize}

\begin{figure*}[ht]
\centering
\includegraphics[width=\linewidth]{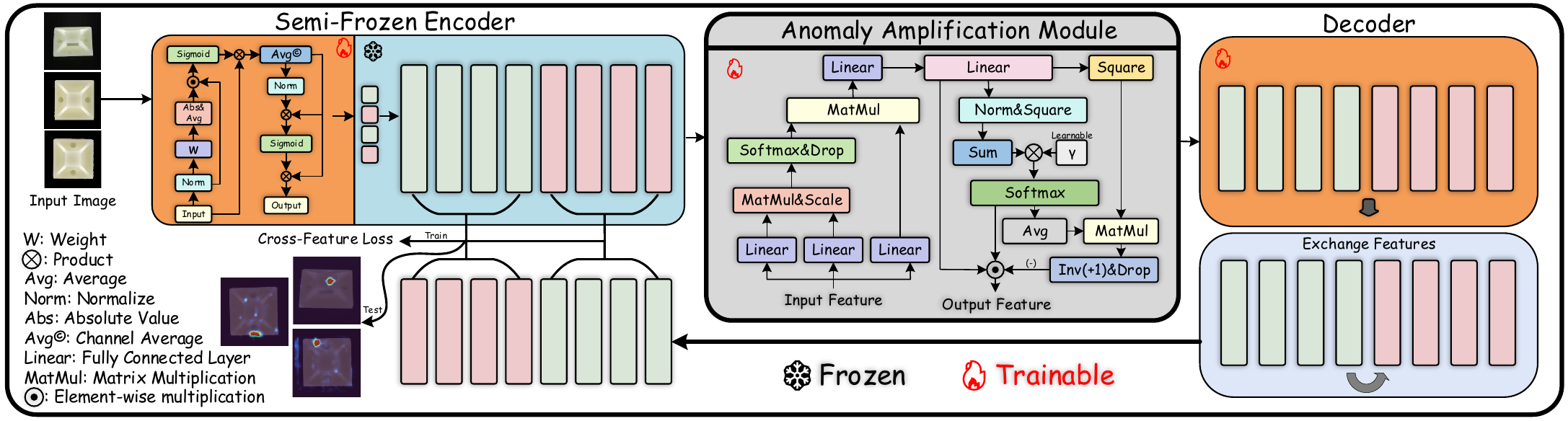}
\caption{\textbf{Framework of our method} that contains three parts: (1) semi-frozen encoder SFE; (2) anomaly amplification module AAM; (3) cross-feature loss CFL. During the training step, the input $x_0$ is put into SFE and AAM to get intermediate features $f_m$. Then the feature $f_m$ is passed through the decoder, and its deep and shallow features are exchanged to compute CFL with features from SFE. During the testing step, $x_0$ is put into the same network to compute the anomaly score.
}
\label{overview}
\end{figure*}

\section{Related Work}
\subsection{Industrial Anomaly Detection}
There has been a growing interest in image anomaly detection \cite{bergmann2019mvtec,wang2024real,liu2021anomaly,dong2024nng,li2021center,yu2024tf}, which plays a crucial role in industrial production. Anomaly detection models are designed to identify whether an image contains anomalies and locate them, which can be categorized into three types: deep feature embedding methods \cite{bergmann2020uninformed,deng2024prioritized,roth2022towards}, reconstruction-based methods \cite{gong2019memorizing,guo2023recontrast,zavrtanik2021reconstruction}, and synthesizing-based methods \cite{li2021cutpaste,zavrtanik2021draem}. Specifically, a current trend in image anomaly detection is to train a unified model rather than separate models for each category. The work \cite{you2022unified} first proposes a unified model that enables multi-class anomaly detection without training separate models for each class. It proposes a unified anomaly detection framework that jointly handles multiple categories using a transformer-based encoder and a dual-branch decoder to model normality across class-shared and class-specific features. It introduces a reconstruction and prediction dual-task setting, enabling efficient multi-class training and significantly improving deployment scalability. RD4AD \cite{deng2022anomaly} introduces a reverse distillation framework that learns a student model to mimic features extracted from a pre-trained one-class teacher, enabling the model to detect anomalies as discrepancies between the two. The method is lightweight and effective, relying solely on normal training data, and achieves strong performance on standard industrial benchmarks. SimpleNet \cite{liu2023simplenet} proposes a lightweight and efficient anomaly detection framework that leverages a single convolutional network with skip connections to directly localize anomalies from normal-only training data. DeSTSeg \cite{zhang2023destseg} presents a segmentation-guided denoising student–teacher framework, where a denoising decoder is trained to reconstruct clean normal features guided by segmentation masks. This approach effectively suppresses irrelevant background noise and enhances the model’s focus on object-centric anomalies, improving pixel-level localization performance. This work \cite{he2024diffusion} proposes a diffusion-based framework for multi-class anomaly detection, where a pre-trained generative diffusion model learns the distribution of normal samples and detects anomalies via reconstruction errors. The method leverages multi-scale denoising and iterative refinement, enabling fine-grained anomaly localization across diverse object categories. MambaAD \cite{hemambaad} explores the use of structured state space models for multi-class unsupervised anomaly detection, leveraging Mamba blocks to efficiently capture long-range dependencies in normal patterns.  Dinomaly \cite{guo2024dinomaly} follows the "less is more" philosophy by directly utilizing frozen DINO ViT features without any fine-tuning, treating anomaly detection as a pure similarity matching task across multi-class categories. 

\subsection{Multi-view Learning}
The methods of multi-view feature fusion are currently widely used. Su et al. \cite{su2015multi} propose a multi-view convolutional neural network (MVCNN) for 3D shape recognition, leveraging multiple 2D views of a 3D object to improve recognition accuracy. Chen et al. \cite{chen2017multi} introduce a multi-view 3D object detection network designed for autonomous driving, utilizing multiple 2D views to enhance the accuracy and robustness of 3D object detection. \cite{pang2022zoom} proposes a mixed-scale triplet network for camouflaged object detection, incorporating multi-view information through zoom-in and zoom-out strategies to enhance detection accuracy in complex scenarios. AIDE \cite{yang2023aide} proposes a vision-driven method for assistive driving perception that leverages multi-view, multi-modal, and multi-tasking information to improve the accuracy and robustness of perception systems. MVSTER \cite{wang2022mvster} introduces an epipolar transformer for efficient multi-view stereo, which utilizes epipolar geometry to enhance the accuracy and computational efficiency of 3D reconstruction from multiple views. He et al. \cite{he2024learning} introduces MVAD, the first dedicated framework for multi-view anomaly detection, featuring the MVAS module which adaptively selects top-K semantically relevant windows for efficient multi-view feature fusion.

\section{Method}

\subsection{Model Overview}
In this section, we detail the overall framework of the proposed method, as illustrated in Figure \ref{overview}. Our approach is built upon a classic encoder-decoder architecture tailored for multi-view multi-class anomaly detection. During training, the decoder learns to reconstruct intermediate features extracted from the encoder by minimizing the Cross-Feature Loss, which is designed to align deep and shallow representations and enhance the model’s ability to capture diverse types of anomalies. In the inference stage, the decoder reconstructs the normal regions well but fails to reconstruct abnormal ones, thereby exposing the deviations that indicate anomalies.

Given an input image $x_0$, we first pass it through a semi-frozen encoder comprises a frozen backbone combined with a pre-encoder prior enhancement mechanism, to get the initial feature $f_i$. Then, $f_i$ is put into the anomaly amplification module to suppress highly similar regions and get the intermediate feature $f_m$, which is the representation with emphasized deviations from normal regions, and can also be written as $f_m = \{f_1, f_2\}$. $f_m$ is then aligned with the encoder features $\{fe_1, fe_2\}$ through a cross-feature loss computation after exchanging deep and shallow features. During inference, the test image $x_0$ is processed through the same semi-frozen encoder and decoder. The pixel-wise reconstruction error between $\{fe_1, fe_2\}$ and its decoder counterpart $f_m = \{f_1, f_2\}$ is used to generate the anomaly heatmap and compute the final anomaly score $S$. This framework enables precise detection and localization of view-specific anomalies across diverse industrial objects.

\subsection{Semi-Frozen Encoder}
To cope with the inconsistency of viewpoint data, we propose a semi-frozen encoder (SFE) that contains a trainable pre-encoder prior mechanism followed by a frozen backbone, providing cross-view shared structural adaptation while preserving its inherent visual representation ability.

Given input images $x_0 \in \mathbb{R}^{B \times C \times H \times W}$, we first pass them into the pre-encoder mechanism to finetune partial layers to enhance adaptability to multi-view images. Specifically, we first perform weighted normalization on an input image $X_{c,h,w} \in \mathbb{R}^{C \times H \times W}$. 
\begin{equation}
\tilde{X}_{c,h,w} = \text{BN}(X) = \gamma_c \cdot \frac{X_{c,h,w} - \mu_c}{\sigma_c}    
\end{equation}
where $\mu_c$ and $\sigma_c$ represent the mean and variance of $X$, $\gamma_c$ serves as the scaling factor of normalization. The weight is defined as,
\begin{equation}
    \alpha_c = \frac{|\gamma_c|}{\sum_{k=1}^{C} |\gamma_k|}
\end{equation}
Therefore, the result after weighting is:

\begin{equation}
    X^{\text{ch}}_{c,h,w} = \sigma\left(\alpha_c \cdot \tilde{X}_{c,h,w}\right) \cdot X_{c,h,w}
\end{equation}
where \( \sigma(\cdot) \) denotes the sigmoid activation function.

Next, we compute the mean of the feature map $X^{\text{ch}}_{c,h,w}$ along the channel dimension to calibrate the variations in input features across different views and improve feature consistency.
\begin{equation}
    \beta_{h,w} = \frac{M_{h,w}}{\sum_{h'=1}^{H} \sum_{w'=1}^{W} M_{h',w'}}
\end{equation}
where $M_{h,w} = \frac{1}{C} \sum_{c=1}^{C} X^{\text{ch}}_{c,h,w}$.

The final output of the prior mechanism is 
\begin{equation}
    X^{\text{prior}}_{c,h,w} = \sigma\left(\beta_{h,w} \cdot X^{\text{ch}}_{c,h,w}\right) \cdot X^{\text{ch}}_{c,h,w}
\end{equation}

Then, the processed features are patched and transformed into a format compatible with the frozen encoder $\varepsilon$. Finally, we get initial feature $f_i$ as follows,
\begin{equation}\label{fi}
    f_i = \varepsilon (Patch(X^{\text{prior}}_{c,h,w}))
\end{equation}

\begin{table*}[htbp]
\centering
\resizebox{1.\linewidth}{!}{
\begin{tabular}{ccccccccc}
\toprule
Method~$\rightarrow$ & RD4AD~\cite{deng2022anomaly} & UniAD~\cite{you2022unified} & SimpleNet~\cite{liu2023simplenet} & DeSTSeg~\cite{zhang2023destseg} & DiAD~\cite{he2024diffusion} & MambaAD~\cite{hemambaad} & Dinomaly\cite{guo2024dinomaly} & \cellcolor{tab_ours} Ours\\
\cline{1-1}
Category~$\downarrow$ & CVPR'22 & NeurlPS'22 & CVPR'23 & CVPR'23 & AAAI'24 & NeurlPS'24 & CVPR'25 & \cellcolor{tab_ours} Ours\\
\hline
audiojack & 76.2/63.2/60.8 & 81.4/76.6/64.9 & 58.4/44.2/50.9 & 81.1/72.6/64.5 & 76.5/54.3/65.7 & 84.2/76.5/67.4 & 86.8/{82.4}/{72.2}  & \cellcolor{tab_ours} \textbf{91.4}/\textbf{87.8}/\textbf{77.6}\\

\cellcolor{tab_others}bottle cap & \cellcolor{tab_others}89.5/86.3/81.0 & \cellcolor{tab_others}92.5/91.7/81.7 & \cellcolor{tab_others}54.1/47.6/60.3 & \cellcolor{tab_others}78.1/74.6/68.1 & \cellcolor{tab_others}91.6/\textbf{94.0}/\textbf{87.9} & \cellcolor{tab_others}\textbf{92.8}/92.0/82.1 & \cellcolor{tab_others}89.9/86.7/81.2  & \cellcolor{tab_ours}90.5/88.4/80.5\\

button battery & 73.3/78.9/76.1 & 75.9/81.6/76.3 & 52.5/60.5/72.4 & 86.7/89.2/83.5 & 80.5/71.3/70.6 & 79.8/85.3/77.8 & 86.6/88.9/\textbf{82.1} & \cellcolor{tab_ours}\textbf{87.5}/\textbf{90.7}/{81.8}\\

\cellcolor{tab_others}end cap & \cellcolor{tab_others}79.8/84.0/77.8 & \cellcolor{tab_others}80.9/86.1/78.0 & \cellcolor{tab_others}51.6/60.8/72.9 & \cellcolor{tab_others}77.9/81.1/77.1 & \cellcolor{tab_others}85.1/83.4/\textbf{84.8} & \cellcolor{tab_others}78.0/82.8/77.2 & \cellcolor{tab_others}\textbf{87.0}/\textbf{87.5}/83.4 & \cellcolor{tab_ours}86.3/86.1/83.7\\

eraser & 90.0/88.7/79.7 & {90.3}/{89.2}/{80.2} & 46.4/39.1/55.8 & 84.6/82.9/71.8 & 80.0/80.0/77.3 & 87.5/86.2/76.1 & 90.3/87.6/78.6 & \cellcolor{tab_ours} \textbf{93.3}/\textbf{91.2}/\textbf{82.7}\\

\cellcolor{tab_others}fire hood & \cellcolor{tab_others}78.3/70.1/64.5 & \cellcolor{tab_others}80.6/74.8/66.4 & \cellcolor{tab_others}58.1/41.9/54.4 & \cellcolor{tab_others}81.7/72.4/67.7 & \cellcolor{tab_others}83.3/\textbf{81.7}/\textbf{80.5} & \cellcolor{tab_others}79.3/72.5/64.8 & \cellcolor{tab_others}83.8/76.2/69.5 &
\cellcolor{tab_ours} \textbf{85.9}/76.4/71.9\\

mint & 65.8/63.1/64.8 & 67.0/66.6/64.6 & 52.4/50.3/63.7 & 58.4/55.8/63.7 & \textbf{76.7}/\textbf{76.7}/\textbf{76.0} & 70.1/70.8/65.5 & 73.1/72.0/67.7 &\cellcolor{tab_ours}76.0/75.3/69.2\\

\cellcolor{tab_others}mounts & \cellcolor{tab_others}88.6/79.9/74.8 & \cellcolor{tab_others}87.6/77.3/77.2 & \cellcolor{tab_others}58.7/48.1/52.4 & \cellcolor{tab_others}74.7/56.5/63.1 & \cellcolor{tab_others}75.3/74.5/\textbf{82.5} & \cellcolor{tab_others}86.8/78.0/73.5 & \cellcolor{tab_others}\textbf{90.4}/\textbf{84.2}/78.0&
\cellcolor{tab_ours}86.7/73.8/77.5\\

pcb & 79.5/85.8/79.7 & 81.0/88.2/79.1 & 54.5/66.0/75.5 & 82.0/88.7/79.6 & 86.0/85.1/85.4 & 89.1/93.7/84.0 & {92.0}/{95.3}/{87.0} &
\cellcolor{tab_ours} \textbf{92.2}/\textbf{95.4}/\textbf{87.7}\\

\cellcolor{tab_others}phone battery & \cellcolor{tab_others}87.5/83.3/77.1 & \cellcolor{tab_others}83.6/80.0/71.6 & \cellcolor{tab_others}51.6/43.8/58.0 & \cellcolor{tab_others}83.3/81.8/72.1 & \cellcolor{tab_others}82.3/77.7/75.9 & \cellcolor{tab_others}90.2/88.9/80.5 & \cellcolor{tab_others}{92.9}/\textbf{91.6}/{82.5}&
\cellcolor{tab_ours} \textbf{93.3}/{91.3}/\textbf{83.5}\\

plastic nut & 80.3/68.0/64.4 & 80.0/69.2/63.7 & 59.2/40.3/51.8 & 83.1/75.4/66.5 & 71.9/58.2/65.6 & 87.1/80.7/70.7 & {88.3}/{81.8}/{74.7} &
\cellcolor{tab_ours} \textbf{92.8}/\textbf{88.2}/\textbf{80.2}\\

\cellcolor{tab_others}plastic plug & \cellcolor{tab_others}81.9/74.3/68.8 & \cellcolor{tab_others}81.4/75.9/67.6 & \cellcolor{tab_others}48.2/38.4/54.6 & \cellcolor{tab_others}71.7/63.1/60.0 & \cellcolor{tab_others}88.7/\textbf{89.2}/\textbf{90.9} & \cellcolor{tab_others}85.7/82.2/72.6 & \cellcolor{tab_others}{90.5}/86.4/78.6 &
\cellcolor{tab_ours} \textbf{92.2}/88.9/80.3\\

porcelain doll & 86.3/76.3/71.5 & 85.1/75.2/69.3 & 66.3/54.5/52.1 & 78.7/66.2/64.3 & 72.6/66.8/65.2 & {88.0}/{82.2}/{74.1} & 85.1/73.3/69.6 &
\cellcolor{tab_ours} \textbf{92.5}/\textbf{87.4}/\textbf{79.5}\\

\cellcolor{tab_others}regulator & \cellcolor{tab_others}66.9/48.8/47.7 & \cellcolor{tab_others}56.9/41.5/44.5 & \cellcolor{tab_others}50.5/29.0/43.9 & \cellcolor{tab_others}79.2/63.5/56.9 & \cellcolor{tab_others}72.1/71.4/\textbf{78.2} & \cellcolor{tab_others}69.7/58.7/50.4 & \cellcolor{tab_others}\textbf{85.2}/\textbf{78.9}/69.8 &
\cellcolor{tab_ours} {83.0}/72.7/61.0\\

rolled strip base & 97.5/98.7/94.7 & 98.7/99.3/96.5 & 59.0/75.7/79.8 & 96.5/98.2/93.0 & 68.4/55.9/56.8 & 98.0/99.0/95.0 & {99.2}/{99.6}/{97.1} &
\cellcolor{tab_ours}\textbf{99.3}/\textbf{99.7}/\textbf{97.2}\\

\cellcolor{tab_others}sim card set & \cellcolor{tab_others}91.6/91.8/84.8 & \cellcolor{tab_others}89.7/90.3/83.2 & \cellcolor{tab_others}63.1/69.7/70.8 & \cellcolor{tab_others}95.5/96.2/{89.2} & \cellcolor{tab_others}72.6/53.7/61.5 & \cellcolor{tab_others}94.4/95.1/87.2 & \cellcolor{tab_others}{95.8}/\textbf{96.3}/88.8 &
\cellcolor{tab_ours} \textbf{96.3}/{95.8}/\textbf{91.7}\\

switch & 84.3/87.2/77.9 & 85.5/88.6/78.4 & 62.2/66.8/68.6 & 90.1/92.8/83.1 & 73.4/49.4/61.2 & 91.7/94.0/85.4 & {97.8}/{98.1}/{93.3} &
\cellcolor{tab_ours}\textbf{98.4}/\textbf{98.7}/\textbf{94.2}\\

\cellcolor{tab_others}tape & \cellcolor{tab_others}96.0/95.1/87.6 & \cellcolor{tab_others}\textbf{97.2}/\textbf{96.2}/\textbf{89.4} & \cellcolor{tab_others}49.9/41.1/54.5 & \cellcolor{tab_others}94.5/93.4/85.9 & \cellcolor{tab_others}73.9/57.8/66.1 & \cellcolor{tab_others}96.8/95.9/89.3 & \cellcolor{tab_others}96.9/95.0/88.8 &
\cellcolor{tab_ours}97.0/94.9/89.7\\

terminalblock & 89.4/89.7/83.1 & 87.5/89.1/81.0 & 59.8/64.7/68.8 & 83.1/86.2/76.6 & 62.1/36.4/47.8 & 96.1/96.8/90.0 & {96.7}/{97.4}/{91.1}&
\cellcolor{tab_ours} \textbf{98.1}/\textbf{98.6}/\textbf{94.3}\\

\cellcolor{tab_others}toothbrush & \cellcolor{tab_others}82.0/83.8/77.2 & \cellcolor{tab_others}78.4/80.1/75.6 & \cellcolor{tab_others}65.9/70.0/70.1 & \cellcolor{tab_others}83.7/85.3/79.0 & \cellcolor{tab_others}\textbf{91.2}/\textbf{93.7}/\textbf{90.9} & \cellcolor{tab_others}85.1/86.2/80.3 & \cellcolor{tab_others}90.4/91.9/83.4&
\cellcolor{tab_ours} 90.4/92.1/83.1\\

toy & 69.4/74.2/75.9 & 68.4/75.1/74.8 & 57.8/64.4/73.4 & 70.3/74.8/75.4 & 66.2/57.3/59.8 & 83.0/87.5/79.6 & 85.6/89.1/81.9 &
\cellcolor{tab_ours} \textbf{87.6}/\textbf{90.9}/\textbf{83.3}\\

\cellcolor{tab_others}toy brick & \cellcolor{tab_others}63.6/56.1/59.0 & \cellcolor{tab_others}\textbf{77.0}/\textbf{71.1}/\textbf{66.2} & \cellcolor{tab_others}58.3/49.7/58.2 & \cellcolor{tab_others}73.2/68.7/{63.3} & \cellcolor{tab_others}68.4/45.3/55.9 & \cellcolor{tab_others}70.5/63.7/61.6 & \cellcolor{tab_others}72.3/65.1/63.4 &
\cellcolor{tab_ours} 77.8/72.3/66.7\\

transistor1 & 91.0/94.0/85.1 & 93.7/95.9/88.9 & 62.2/69.2/72.1 & 90.2/92.1/84.6 & 73.1/63.1/62.7 & 94.4/96.0/89.0 & \textbf{97.4}/\textbf{98.2}/\textbf{93.1} &
\cellcolor{tab_ours}\textbf{97.4}/\textbf{98.2}/92.9\\

\cellcolor{tab_others}u block & \cellcolor{tab_others}89.5/85.0/74.2 & \cellcolor{tab_others}88.8/84.2/75.5 & \cellcolor{tab_others}62.4/48.4/51.8 & \cellcolor{tab_others}80.1/73.9/64.3 & \cellcolor{tab_others}75.2/68.4/67.9 & \cellcolor{tab_others}89.7/{85.7}/{75.3} & \cellcolor{tab_others}{89.9}/84.0/75.2 &
\cellcolor{tab_ours} \textbf{94.9}/\textbf{92.6}/\textbf{83.9}\\

usb & 84.9/84.3/75.1 & 78.7/79.4/69.1 & 57.0/55.3/62.9 & 87.8/88.0/78.3 & 58.9/37.4/45.7 & {92.0}/{92.2}/{84.5} & {92.0}/91.6/83.3 &
\cellcolor{tab_ours} \textbf{94.5}/\textbf{94.0}/\textbf{86.8}\\

\cellcolor{tab_others}usb adaptor & \cellcolor{tab_others}71.1/61.4/62.2 & \cellcolor{tab_others}76.8/71.3/64.9 & \cellcolor{tab_others}47.5/38.4/56.5 & \cellcolor{tab_others}80.1/{74.9}/67.4 & \cellcolor{tab_others}76.9/60.2/67.2 & \cellcolor{tab_others}79.4/76.0/66.3 & \cellcolor{tab_others}{81.5}/74.5/{69.4} &
\cellcolor{tab_ours} \textbf{86.3}/\textbf{81.1}/\textbf{74.0}\\

vcpill & 85.1/80.3/72.4 & 87.1/84.0/74.7 & 59.0/48.7/56.4 & 83.8/81.5/69.9 & 64.1/40.4/56.2 & 88.3/87.7/77.4 & {92.0}/{91.2}/{82.0} &
\cellcolor{tab_ours} \textbf{93.9}/\textbf{93.5}/\textbf{85.0}\\

\cellcolor{tab_others}wooden beads & \cellcolor{tab_others}81.2/78.9/70.9 & \cellcolor{tab_others}78.4/77.2/67.8 & \cellcolor{tab_others}55.1/52.0/60.2 & \cellcolor{tab_others}82.4/78.5/73.0 & \cellcolor{tab_others}62.1/56.4/65.9 & \cellcolor{tab_others}82.5/81.7/71.8 & \cellcolor{tab_others}{87.3}/{85.8}/{77.4} &
\cellcolor{tab_ours} \textbf{90.0}/\textbf{89.1}/\textbf{80.7}\\

woodstick & 76.9/61.2/58.1 & 80.8/72.6/63.6 & 58.2/35.6/45.2 & 80.4/69.2/60.3 & 74.1/66.0/62.1 & 80.4/69.0/63.4 & \textbf{84.0}/\textbf{73.3}/\textbf{65.6}&
\cellcolor{tab_ours}\textbf{84.0}/72.0/65.7\\

\cellcolor{tab_others}zipper & \cellcolor{tab_others}95.3/97.2/91.2 & \cellcolor{tab_others}98.2/98.9/95.3 & \cellcolor{tab_others}77.2/86.7/77.6 & \cellcolor{tab_others}96.9/98.1/93.5 & \cellcolor{tab_others}86.0/87.0/84.0 & \cellcolor{tab_others}\textbf{99.2}/\textbf{99.6}/\textbf{96.9} & \cellcolor{tab_others}99.1/99.5/96.5&
\cellcolor{tab_ours} 99.1/99.5/96.5\\
\midrule
Mean & 82.4/79.0/73.9 & 83.0/80.9/74.3 & 57.2/53.4/61.5 & 82.3/79.2/73.2 & 75.6/66.4/69.9 & 86.3/84.6/77.0 & {89.3}/{86.8}/{80.2} &
\cellcolor{tab_ours} \textbf{91.0}/\textbf{88.6}/\textbf{82.1}\\
\bottomrule
\end{tabular}
}
\vspace{5pt}
\caption{Per-class performance on \textbf{Real-IAD} dataset for multi-class anomaly detection with AUROC/AP/$F_1$-max metrics.}
\label{tab:realiadsp}
\end{table*}

\subsection{Anomaly Amplification Module}
To better emphasize the representation of anomaly regions, we introduce the anomaly amplification module (AAM), which captures global token-to-token contextual relationships, and focuses on suppressing tokens that are highly similar to those likely representing normal regions by computing normalized similarity and inversely weighting feature responses. It not only helps the model aggregate relevant information from different parts of the image, but also suppresses dominant normal patterns while amplifying semantically deviant or rare tokens, which often correspond to anomalies.

Given $f_i \in \mathbb{R}^{B \times N \times D}$ obtained by equation (\ref{fi}), where $B$, $N$, $D$ represent batch size, token number, token dimension, respectively. The anomaly amplification module first models global token-level context and passes it into a similarity suppression mechanism. Then we compute a soft attention distribution $\Pi$ based on token norms and a learnable temperature $\gamma$ and suppress the dominant normal patterns by applying inverse weighting. The final output emphasizes the semantic deviations, highlighting the anomalies while suppressing redundant normal features.

Specifically, we first compute the query, key, and value matrices:
\begin{equation}
    Q = f_iW^Q, \quad K = f_iW^K, \quad V = f_iW^V
\end{equation}
where $Q, K, V \in \mathbb{R}^{B \times h \times N \times d_k}$ are linear mapping weights, $d_k$ is the dimension of attention head. The
weighted sum of the value is calculated as follows,
\begin{equation}
    F = W^F(\text{Softmax}\left( \frac{QK^\top}{\sqrt{d_k}} \right)V)
\end{equation}

where $W^F$ is linear mapping weight. We normalize the features $F$ along the token dimension:
\begin{equation}
    \hat{F} = \text{Normalize}(F, \text{dim}=N)
\end{equation}

We then compute the token similarity scores:
\begin{equation}
    Sim = \sum_{j=1}^{N} \|\hat{F}_{j}\|^2 \cdot \gamma 
\end{equation}

where $\quad \gamma \in \mathbb{R}^{h \times 1} $ is a learnable temperature parameter.

We define a soft attention distribution, 
\begin{equation}
    \Pi = \text{Softmax}(Sim) \in \mathbb{R}^{B \times h \times N}
\end{equation}

We compute a suppression-based attention factor,
\begin{equation}
    \text{Att}= \frac{1}{1 + \left( \Pi^\top \cdot F^2 \right)}
\end{equation}

Therefore, the final output of anomaly amplification module is,
\begin{equation}\label{fm}
    f_m = W^{out} (- (F \cdot \Pi) \cdot \text{Att})
\end{equation}

where $W^F$ is linear mapping weight.

The AAM enhances semantic deviations by modeling global token interactions and suppressing dominant patterns, thereby emphasizing anomaly-relevant tokens in a multi-view setting.

\subsection{Cross-Feature Loss}
To capture anomalies manifested at different semantic levels, cross-feature loss is designed to align shallow encoder features with deep decoder features and vice versa, which  enables the model to detect both low-level texture deviations and high-level structural inconsistencies, thereby improving its robustness and sensitivity to multi-scale anomalies.

Given the features \{$f_{e1},f_{e2}$\} from semi-frozen encoder and features $f_m = \{ f_1,f_2\}$ from equation (\ref{fm}), we align the features by exchanging shallow and deep layers, that is to say, we align $f_{e1}$ with $f_2$ and $f_{e2}$ with $f_1$. We define the similarity as follows, for a feature pair $(z_1,z_2)$
\begin{equation}
    \text{Score} = 1-cos(z_1,z_2)
\end{equation}

Let $h$ be the similarity score ranked exactly at the top 10\%, we select the set consisting of the top 10\% highest similarity scores.
\begin{equation}
    \mathcal{I} = \left\{ i \mid \text{Score}_i \geq h \right\}
\end{equation}

The cross-feature loss is then computed as:
\begin{align}
\mathcal{L}_{\text{cross}} = \frac{1}{2} \bigg( 
& \frac{1}{|\mathcal{I}|} \sum_{i \in \mathcal{I}}  \text{Score}(f_{e1}, f_2) \notag \\
+ & \frac{1}{|\mathcal{I}|} \sum_{i \in \mathcal{I}}  \text{Score}(f_{e2}, f_1) 
\bigg)
\end{align}

\section{Experiments}
\subsection{Datasets and Metrics}
Real-IAD (Real-world Industrial Anomaly Detection) \cite{wang2024real} is a large-scale, multi-view industrial anomaly detection dataset, designed to provide a challenging benchmark platform for the field of Industrial Anomaly Detection (IAD). It contains images of 30 different object categories, with a total of 36,465 normal images in the training set and 114,585 images in the test set (63,256 normal, 51,329 anomalous). The operators capture images from five different camera angles (including top and four symmetrical views) for each object, reflecting the multi-view nature of real-world industrial scenarios. The images are captured with professional cameras, featuring high resolution and clear details, and include defects with varying sizes and proportions.

Following the previous work \cite{hemambaad,guo2024dinomaly}, we choose seven evaluation metrics. For image-level anomaly detection, we use the Area Under the Receiver Operator Curve (AUROC), Average Precision (AP), and the $F_1$ score under optimal threshold ($F_1$-max). And for pixel-level anomaly localization, we use AUROC, AP, $F_1$-max and the Area Under the Per-Region-Overlap (AUPRO).

\begin{table*}[htbp!]
\centering
\resizebox{1.\linewidth}{!}{
\begin{tabular}{ccccccccc}
\toprule
Method~$\rightarrow$ & RD4AD~\cite{deng2022anomaly} & UniAD~\cite{you2022unified} & SimpleNet~\cite{liu2023simplenet} & DeSTSeg~\cite{zhang2023destseg} & DiAD~\cite{he2024diffusion} & MambaAD~\cite{hemambaad} & Dinomaly~\cite{guo2024dinomaly}& \cellcolor{tab_ours}Ours \\
\cline{1-1}
Category~$\downarrow$ & CVPR'22 & NeurlPS'22 & CVPR'23 & CVPR'23 & AAAI'24 & NeurlPS'24 & CVPR'25&\cellcolor{tab_ours} Ours \\
\hline
audiojack & 96.6/12.8/22.1/79.6 & 97.6/20.0/31.0/83.7 & 74.4/ 0.9/ 4.8/38.0 & 95.5/25.4/31.9/52.6 & 91.6/ 1.0/ 3.9/63.3 & 97.7/21.6/29.5/83.9 & {98.7}/{48.1}/\textbf{54.5}/{91.7} &
\cellcolor{tab_ours} \textbf{99.0}/\textbf{51.8}/{54.2}/\textbf{93.5}\\

\cellcolor{tab_others}bottle cap & \cellcolor{tab_others}99.5/18.9/29.9/95.7 & \cellcolor{tab_others}99.5/19.4/29.6/96.0 & \cellcolor{tab_others}85.3/ 2.3/ 5.7/45.1 & \cellcolor{tab_others}94.5/25.3/31.1/25.3 & \cellcolor{tab_others}94.6/ 4.9/11.4/73.0 & \textbf{99.7}/30.6/34.6/97.2 & \cellcolor{tab_others}\textbf{99.7}/{32.4}/{36.7}/{98.1}&
\cellcolor{tab_ours}\textbf{99.7}/\textbf{35.0}/\textbf{41.3}/\textbf{98.2}\\

button battery& 97.6/33.8/37.8/86.5 & 96.7/28.5/34.4/77.5 & 75.9/ 3.2/ 6.6/40.5 & 98.3/\textbf{63.9}/\textbf{60.4}/36.9 & 84.1/ 1.4/ 5.3/66.9 & 98.1/46.7/49.5/86.2 & \textbf{99.1}/46.9/56.7/{92.9} &
\cellcolor{tab_ours}\textbf{99.1}/60.9/57.7/\textbf{93.1}\\

\cellcolor{tab_others}end cap & \cellcolor{tab_others}96.7/12.5/22.5/89.2 & \cellcolor{tab_others}95.8/ 8.8/17.4/85.4 & \cellcolor{tab_others}63.1/ 0.5/ 2.8/25.7 & \cellcolor{tab_others}89.6/14.4/22.7/29.5 & \cellcolor{tab_others}81.3/ 2.0/ 6.9/38.2 & 97.0/12.0/19.6/89.4 & \cellcolor{tab_others}\textbf{99.1}/\textbf{26.2}/\textbf{32.9}/\textbf{96.0} &
\cellcolor{tab_ours}{98.8}/17.5/28.4/{95.3}\\

eraser & \textbf{99.5}/30.8/36.7/96.0 & 99.3/24.4/30.9/94.1 & 80.6/ 2.7/ 7.1/42.8 & 95.8/52.7/53.9/46.7 & 91.1/ 7.7/15.4/67.5 & 99.2/30.2/38.3/93.7 & {99.5}/{39.6}/{43.3}/{96.4}&
\cellcolor{tab_ours}\textbf{99.8}/\textbf{41.8}/\textbf{46.8}/\textbf{98.4}\\

\cellcolor{tab_others}fire hood & \cellcolor{tab_others}98.9/27.7/35.2/87.9 & \cellcolor{tab_others}98.6/23.4/32.2/85.3 & \cellcolor{tab_others}70.5/ 0.3/ 2.2/25.3 & \cellcolor{tab_others}97.3/27.1/35.3/34.7 & \cellcolor{tab_others}91.8/ 3.2/ 9.2/66.7 & 98.7/25.1/31.3/86.3 & \cellcolor{tab_others}{99.3}/\textbf{38.4}/\textbf{42.7}/{93.0}&
\cellcolor{tab_ours} \textbf{99.4}/30.4/38.7/\textbf{95.0}\\

mint & 95.0/11.7/23.0/72.3 & 94.4/ 7.7/18.1/62.3 & 79.9/ 0.9/ 3.6/43.3 & 84.1/10.3/22.4/ 9.9 & 91.1/ 5.7/11.6/64.2 & 96.5/15.9/27.0/72.6 & {96.9}/{22.0}/{32.5}/{77.6}&
\cellcolor{tab_ours}\textbf{97.8}/\textbf{28.0}/\textbf{37.7}/\textbf{82.4}\\

\cellcolor{tab_others}mounts & \cellcolor{tab_others}99.3/30.6/37.1/94.9 & \cellcolor{tab_others}\textbf{99.4}/28.0/32.8/95.2 & \cellcolor{tab_others}80.5/ 2.2/ 6.8/46.1 & \cellcolor{tab_others}94.2/30.0/41.3/43.3 & \cellcolor{tab_others}84.3/ 0.4/ 1.1/48.8 & 99.2/31.4/35.4/93.5 & \cellcolor{tab_others}{99.4}/{39.9}/{44.3}/{95.6}&
\cellcolor{tab_ours} \textbf{99.6}/\textbf{40.5}/\textbf{44.3}/\textbf{97.4}\\

pcb & 97.5/15.8/24.3/88.3 & 97.0/18.5/28.1/81.6 & 78.0/ 1.4/ 4.3/41.3 & 97.2/37.1/40.4/48.8 & 92.0/ 3.7/ 7.4/66.5 & 99.2/46.3/50.4/93.1 & {99.3}/{55.0}/\textbf{56.3}/\textbf{95.7}&
\cellcolor{tab_ours}\textbf{99.4}/\textbf{55.6}/{52.7}/\textbf{95.7}\\

\cellcolor{tab_others}phone battery & \cellcolor{tab_others}77.3/22.6/31.7/94.5 & \cellcolor{tab_others}85.5/11.2/21.6/88.5 & \cellcolor{tab_others}43.4/ 0.1/ 0.9/11.8 & \cellcolor{tab_others}79.5/25.6/33.8/39.5 & \cellcolor{tab_others}96.8/ 5.3/11.4/85.4 & 99.4/36.3/41.3/95.3 & \cellcolor{tab_others}\textbf{99.7}/\textbf{51.6}/\textbf{54.2}/\textbf{96.8}&
\cellcolor{tab_ours} \textbf{99.7}/{45.1}/{47.7}/{96.7}\\

plastic nut& 98.8/21.1/29.6/91.0 &98.4/20.6/27.1/88.9 &77.4/ 0.6/ 3.6/41.5 &96.5/44.8/45.7/38.4 &81.1/ 0.4/ 3.4/38.6 &99.4/33.1/37.3/96.1 & \textbf{99.7}/\textbf{41.0}/\textbf{45.0}/{97.4}&
\cellcolor{tab_ours} \textbf{99.8}/{39.5}/{44.6}/\textbf{98.1}\\

\cellcolor{tab_others}plastic plug& \cellcolor{tab_others}99.1/20.5/28.4/94.9 &\cellcolor{tab_others}98.6/17.4/26.1/90.3 &\cellcolor{tab_others}78.6/ 0.7/ 1.9/38.8 &\cellcolor{tab_others}91.9/20.1/27.3/21.0 &\cellcolor{tab_others}92.9/ 8.7/15.0/66.1 &99.0/24.2/31.7/91.5 & \cellcolor{tab_others}{99.4}/\textbf{31.7}/{37.2}/{96.4}&
\cellcolor{tab_ours} \textbf{99.6}/{31.3}/\textbf{38.2}/\textbf{98.0}\\

porcelain doll& 99.2/24.8/34.6/95.7 &98.7/14.1/24.5/93.2 &81.8/ 2.0/ 6.4/47.0 &93.1/35.9/40.3/24.8 &93.1/ 1.4/ 4.8/70.4 &99.2/{31.3}/\textbf{36.6}/95.4 & {99.3}/27.9/33.9/{96.0}&
\cellcolor{tab_ours} \textbf{99.6}/\textbf{38.1}/\textbf{43.5}/\textbf{98.1}\\

\cellcolor{tab_others}regulator& \cellcolor{tab_others}98.0/7.8/16.1/88.6 &\cellcolor{tab_others}95.5/9.1/17.4/76.1 &\cellcolor{tab_others}76.6/0.1/0.6/38.1 &\cellcolor{tab_others}88.8/18.9/23.6/17.5 &\cellcolor{tab_others}84.2/0.4/1.5/44.4 &97.6/20.6/29.8/87.0 & \cellcolor{tab_others}{99.3}/\textbf{42.2}/\textbf{48.9}/{95.6}&
\cellcolor{tab_ours} \textbf{99.4}/40.9/47.7/\textbf{96.2}\\

rolled strip base& {99.7}/31.4/39.9/98.4 &99.6/20.7/32.2/97.8 &80.5/ 1.7/ 5.1/52.1 &99.2/{48.7}/{50.1}/55.5 &87.7/ 0.6/ 3.2/63.4 &99.7/37.4/42.5/98.8 & {99.7}/41.6/45.5/{98.5}&
\cellcolor{tab_ours} \textbf{99.9}/\textbf{53.2}/\textbf{57.5}/\textbf{99.2}\\

\cellcolor{tab_others}sim card set& \cellcolor{tab_others}98.5/40.2/44.2/89.5 &\cellcolor{tab_others}97.9/31.6/39.8/85.0 &\cellcolor{tab_others}71.0/ 6.8/14.3/30.8 &\cellcolor{tab_others}{99.1}/\textbf{65.5}/\textbf{62.1}/73.9 &\cellcolor{tab_others}89.9/ 1.7/ 5.8/60.4 &98.8/51.1/50.6/89.4 & \cellcolor{tab_others}99.0/52.1/52.9/{90.9}&
\cellcolor{tab_ours} \textbf{99.2}/41.6/49.2/\textbf{93.6}\\

switch& 94.4/18.9/26.6/90.9 &98.1/33.8/40.6/90.7 &71.7/ 3.7/ 9.3/44.2 &97.4/57.6/55.6/44.7 &90.5/ 1.4/ 5.3/64.2 &\textbf{98.2}/39.9/45.4/92.9 & 96.7/\textbf{62.3}/\textbf{63.6}/\textbf{95.9}&
\cellcolor{tab_ours}97.1/58.3/58.6/\textbf{95.9}\\

\cellcolor{tab_others}tape& \cellcolor{tab_others}99.7/42.4/47.8/98.4 &\cellcolor{tab_others}99.7/29.2/36.9/97.5 &\cellcolor{tab_others}77.5/ 1.2/ 3.9/41.4 &\cellcolor{tab_others}99.0/61.7/57.6/48.2 &\cellcolor{tab_others}81.7/ 0.4/ 2.7/47.3 &\textbf{99.8}/47.1/48.2/98.0 & \cellcolor{tab_others}\textbf{99.8}/\textbf{54.0}/\textbf{55.8}/{98.8}&
\cellcolor{tab_ours}\textbf{99.8}/{53.5}/{54.9}/\textbf{99.0}\\

terminalblock& 99.5/27.4/35.8/97.6 &99.2/23.1/30.5/94.4 &87.0/ 0.8/ 3.6/54.8 &96.6/40.6/44.1/34.8 &75.5/ 0.1/ 1.1/38.5 &\textbf{99.8}/35.3/39.7/98.2 & \textbf{99.8}/{48.0}/{50.7}/{98.8}&
\cellcolor{tab_ours} \textbf{99.8}/\textbf{48.3}/\textbf{50.8}/\textbf{99.1}\\

\cellcolor{tab_others}toothbrush& \cellcolor{tab_others}96.9/26.1/34.2/88.7 &\cellcolor{tab_others}95.7/16.4/25.3/84.3 &\cellcolor{tab_others}84.7/ 7.2/14.8/52.6 &\cellcolor{tab_others}94.3/30.0/37.3/42.8&\cellcolor{tab_others}82.0/ 1.9/ 6.6/54.5&\textbf{97.5}/27.8/36.7/\textbf{91.4} & \cellcolor{tab_others}96.9/{38.3}/{43.9}/90.4&
\cellcolor{tab_ours}97.2/\textbf{39.5}/\textbf{45.2}/91.3\\

toy & 95.2/ 5.1/12.8/82.3 & 93.4/ 4.6/12.4/70.5 &67.7/ 0.1/ 0.4/25.0 &86.3/ 8.1/15.9/16.4 &82.1/ 1.1/ 4.2/50.3 &\textbf{96.0}/16.4/25.8/86.3 & 94.9/{22.5}/{32.1}/{91.0}&
\cellcolor{tab_ours}95.5/\textbf{24.2}/\textbf{33.3}/\textbf{91.5}\\

\cellcolor{tab_others}toy brick& \cellcolor{tab_others}96.4/16.0/24.6/75.3 &\cellcolor{tab_others}{97.4}/17.1/27.6/\textbf{81.3} &\cellcolor{tab_others}86.5/ 5.2/11.1/56.3 &\cellcolor{tab_others}94.7/24.6/30.8/45.5 &\cellcolor{tab_others}93.5/ 3.1/ 8.1/66.4 &96.6/18.0/25.8/74.7 & \cellcolor{tab_others}96.8/{27.9}/{34.0}/76.6&
\cellcolor{tab_ours} \textbf{97.7}/\textbf{31.0}/\textbf{37.6}/\textbf{83.6}\\

transistor1& 99.1/29.6/35.5/95.1 &98.9/25.6/33.2/94.3 &71.7/ 5.1/11.3/35.3 &97.3/43.8/44.5/45.4 &88.6/ 7.2/15.3/58.1 &99.4/39.4/40.0/96.5 & \textbf{99.6}/{53.5}/{53.3}/\textbf{97.8}&
\cellcolor{tab_ours} \textbf{99.6}/\textbf{55.3}/\textbf{54.2}/\textbf{97.8}\\

\cellcolor{tab_others}u block& \cellcolor{tab_others}99.6/40.5/45.2/96.9 &\cellcolor{tab_others}99.3/22.3/29.6/94.3 &\cellcolor{tab_others}76.2/ 4.8/12.2/34.0 &\cellcolor{tab_others}96.9/\textbf{57.1}/\textbf{55.7}/38.5 &\cellcolor{tab_others}88.8/ 1.6/ 5.4/54.2 &{99.5}/37.8/46.1/95.4 & \cellcolor{tab_others}{99.5}/41.8/45.6/{96.8}&
\cellcolor{tab_ours} \textbf{99.8}/50.0/55.2/\textbf{98.5}\\

usb& 98.1/26.4/35.2/91.0 &97.9/20.6/31.7/85.3 &81.1/ 1.5/ 4.9/52.4 &98.4/42.2/47.7/57.1 &78.0/ 1.0/ 3.1/28.0 &{99.2}/39.1/44.4/95.2 & {99.2}/\textbf{45.0}/\textbf{48.7}/{97.5}&
\cellcolor{tab_ours} \textbf{99.4}/{42.9}/{48.0}/\textbf{98.0}\\

\cellcolor{tab_others}usb adaptor& \cellcolor{tab_others}94.5/ 9.8/17.9/73.1 &\cellcolor{tab_others}96.6/10.5/19.0/78.4 &\cellcolor{tab_others}67.9/ 0.2/ 1.3/28.9 &\cellcolor{tab_others}94.9/{25.5}/{34.9}/36.4 &\cellcolor{tab_others}94.0/ 2.3/ 6.6/75.5 &97.3/15.3/22.6/82.5 & \cellcolor{tab_others}{98.7}/23.7/32.7/{91.0}&
\cellcolor{tab_ours} \textbf{99.0}/\textbf{27.5}/\textbf{36.9}/\textbf{92.9}\\

vcpill& 98.3/43.1/48.6/88.7 &99.1/40.7/43.0/91.3 &68.2/ 1.1/ 3.3/22.0 &97.1/64.7/62.3/42.3 &90.2/ 1.3/ 5.2/60.8 &98.7/50.2/54.5/89.3 & {99.1}/{66.4}/{66.7}/{93.7}&
\cellcolor{tab_ours}\textbf{99.4}/\textbf{70.9}/\textbf{68.9}/\textbf{95.7}\\

\cellcolor{tab_others}wooden beads&	\cellcolor{tab_others}98.0/27.1/34.7/85.7 &\cellcolor{tab_others}97.6/16.5/23.6/84.6 &\cellcolor{tab_others}68.1/ \;2.4/ \;6.0/28.3 &\cellcolor{tab_others}94.7/38.9/42.9/39.4 &\cellcolor{tab_others}85.0/ \;1.1/ \;4.7/45.6 &\cellcolor{tab_others}98.0/32.6/39.8/84.5 & \cellcolor{tab_others}{99.1}/{45.8}/{50.1}/{90.5}&
\cellcolor{tab_ours} \textbf{99.3}/\textbf{47.0}/\textbf{51.0}/\textbf{93.3}\\

woodstick& 97.8/30.7/38.4/85.0 &94.0/36.2/44.3/77.2 &76.1/ 1.4/ 6.0/32.0 &97.9/\textbf{60.3}/\textbf{60.0}/51.0 &90.9/ 2.6/ 8.0/60.7 &97.7/40.1/44.9/82.7 & {99.0}/50.9/52.1/{90.4}&
\cellcolor{tab_ours}\textbf{99.2}/47.0/50.3/\textbf{92.3}\\

\cellcolor{tab_others}zipper& \cellcolor{tab_others}99.1/44.7/50.2/96.3 &\cellcolor{tab_others}98.4/32.5/36.1/95.1 &\cellcolor{tab_others}89.9/23.3/31.2/55.5 &\cellcolor{tab_others}98.2/35.3/39.0/78.5 &\cellcolor{tab_others}90.2/12.5/18.8/53.5 &99.3/58.2/61.3/97.6 & \cellcolor{tab_others}{99.3}/{67.2}/{66.5}/\textbf{97.8}&
\cellcolor{tab_ours} \textbf{99.4}/\textbf{70.7}/\textbf{67.9}/97.4\\

\hline
Mean& 97.3/25.0/32.7/89.6 &97.3/21.1/29.2/86.7 &75.7/ 2.8/ 6.5/39.0 &94.6/37.9/41.7/40.6 &88.0/ 2.9/ 7.1/58.1 &98.5/33.0/38.7/90.5 & {98.8}/{42.8}/{47.1}/{93.9}&
\cellcolor{tab_ours} \textbf{99.1}/\textbf{43.9}/\textbf{48.2}/\textbf{95.2}\\

\bottomrule
\end{tabular}
}
\vspace{5pt}
\caption{Per-class performance on \textbf{Real-IAD} dataset for multi-class anomaly localization with AUROC/AP/$F_1$-max/AUPRO metrics.}
\label{tab:realiadpx}
\end{table*}

\begin{table*}[!h]
  \centering
    \tiny
  
   \resizebox{0.9\linewidth}{!}{
    \begin{tabular}{cccccccccc}
    \toprule
     \multirow{2}[2]{*}{SFE} & \multirow{2}[2]{*}{AAM} & \multirow{2}[2]{*}{CFL} & \multicolumn{3}{c}{Image-level} & \multicolumn{4}{c}{Pixel-level} \\
\cmidrule(l){4-6} \cmidrule(l){7-10} 
 & & & \multicolumn{1}{c}{AUROC} & \multicolumn{1}{c}{AP} & \multicolumn{1}{c}{$F_1$-max} & \multicolumn{1}{c}{AUROC} & \multicolumn{1}{c}{AP} & \multicolumn{1}{c}{$F_1$-max} & \multicolumn{1}{c}{AUPRO} \\ \midrule
       &    &    &  84.8&81.4&76.0&97.7&32.5&39.4&91.4\\
    \checkmark  &    &    & 85.9 & 82.9 &76.8  & 98.2 & 35.9 & 42.5  & 92.2 \\
     & \checkmark  &    & 87.9 & 85.1 &  78.6 & 98.6 & 39.4 & 44.9  & 92.0 \\
     &    & \checkmark  & 84.9 & 81.4 &  75.5 & 98.0 & 32.9 &  40.0 &  91.7\\
\checkmark    &   &  \checkmark   & 90.4 & 87.7 &  81.4 &  98.9& 45.0 &  48.9 & 94.5 \\

   & \checkmark  & \checkmark  &  89.7  &  87.2 &  80.4 &  98.9  & 43.9  & 48.0     &   94.2  \\
 \checkmark  & \checkmark  &    & 90.1 & 87.5 & 80.9  & 98.9 & 44.7 & 48.9  & 94.7 \\
 \checkmark  & \checkmark  & \checkmark  & 91.0 & 88.6 & 82.1 & 99.1 & 43.9 & 48.2 & 95.2 \\ \bottomrule
\end{tabular}
}
\vspace{5pt}
\caption{Ablations of proposed components on Real-IAD (\%). SFE, AAM and CFL represent Semi-Frozen Encoder, Anomaly Amplification Module and Cross-Feature Loss, respectively.}
\label{tab:abiad}
\end{table*}

\begin{figure*}[ht]
\centering
\includegraphics[width=\linewidth]{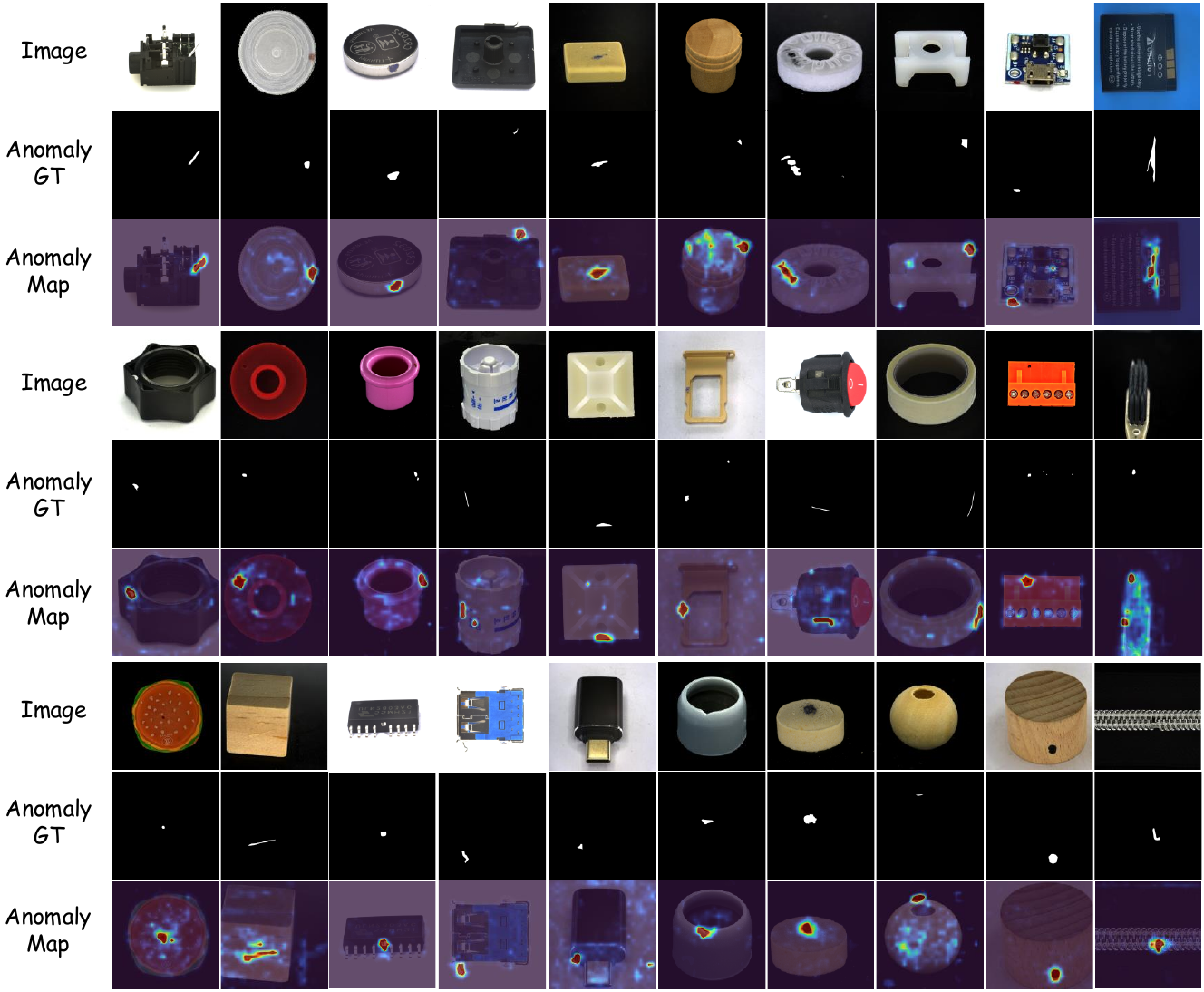}
\caption{\textbf{Visualization of Anomaly Map on Real-IAD.} All samples are randomly chosen. The anomaly map shows that our method is capable of precisely localizing anomalies across diverse object types,
even when the defects are small or subtle.
}
\label{vis}
\end{figure*}

\begin{table*}[htbp]
  \centering
  \resizebox{1\linewidth}{!}{
  \begin{tabular}{ccc|>{\columncolor{tab_ours}}c|cc|>{\columncolor{tab_ours}}c}
  \toprule
        & \multicolumn{3}{c|}{\textbf{Image-level separate-class} AUROC/AP/$F_1$-max} & \multicolumn{3}{c}{\textbf{Pixel-level separate-class} AUROC/AP/$F_1$-max/PRO} \\
  \cmidrule{2-7}   Method & {UniAD~\cite{you2022unified}} & {MVAD~\cite{he2024learning}} & Ours & {UniAD~\cite{you2022unified}} & {MVAD~\cite{he2024learning}} & Ours \\
  \midrule
audiojack & 78.7/60.8/64.0 & 86.9/82.5/70.6 & 91.4/87.4/78.2 & 97.2/7.8/14.4/83.9 & 98.8/36.6/44.4/90.6 & 99.1/47.4/52.8/94.6 \\
bottle\_cap & 85.6/82.6/74.9 & 95.6/95.4/87.0 & 90.9/88.8/79.7 & 99.2/21.4/30.9/93.8 & 99.7/36.2/39.5/97.4 & 98.5/14.7/25.7/85.8 \\
button\_battery & 65.9/71.9/74.6 & 90.8/92.1/86.8 & 88.5/90.2/83.1 & 93.7/13.6/20.7/70.3 & 98.8/46.6/45.7/88.5 & 91.5/38.2/52.8/91.5 \\
end\_cap & 80.6/84.4/78.3 & 85.8/88.8/81.8 & 84.7/85.4/77.3 & 96.7/7.6/15.2/88.3 & 98.1/12.7/21.5/92.6 & 94.0/15.5/26.3/93.0 \\
eraser & 87.9/82.4/75.5 & 91.2/89.2/79.7 & 94.8/93.6/84.7 & 99.0/17.7/24.6/92.5 & 99.2/30.7/35.9/92.3 & 98.3/45.5/48.5/98.3 \\
fire\_hood & 79.0/72.3/65.0 & 84.6/77.2/69.8 & 85.9/80.1/71.1 & 98.5/22.7/31.2/84.6 & 99.1/28.5/37.1/90.1 & 95.2/41.8/47.1/95.2 \\
mint & 64.5/63.8/63.9 & 79.5/80.7/70.8 & 75.4/74.9/69.7 & 94.3/6.1/15.8/59.3 & 98.5/19.7/27.9/83.1 & 79.6/24.1/34.1/79.6 \\
mounts & 84.1/71.2/71.0 & 87.9/75.6/77.3 & 85.6/72.4/74.0 & 99.4/27.9/35.3/95.4 & 99.0/30.2/34.0/90.3 & 96.0/39.9/44.5/96.0 \\
pcb & 84.0/89.2/81.9 & 91.3/94.6/86.2 & 88.8/93.3/84.3 & 96.6/4.5/9.5/79.6 & 99.4/50.4/53.9/94.3 & 93.6/46.4/49.4/93.6 \\
phone\_battery & 83.7/75.9/73.5 & 92.5/90.7/82.2 & 91.4/87.9/81.1 & 97.9/8.0/14.1/87.5 & 99.2/38.6/46.1/93.7 & 95.2/55.6/53.3/95.2 \\
plastic\_nut & 78.7/64.7/62.0 & 91.3/85.7/77.5 & 93.6/88.7/81.6 & 98.6/16.4/23.9/90.1 & 99.6/32.5/35.1/96.8 & 98.3/41.1/43.4/98.3 \\
plastic\_plug & 70.7/59.9/61.1 & 89.7/87.2/76.0 & 91.3/87.6/78.4 & 97.9/10.3/18.2/85.9 & 99.0/24.4/29.8/92.0 & 97.9/32.0/37.2/97.9 \\
porcelain\_doll & 68.3/53.9/53.6 & 87.8/81.5/72.9 & 91.1/85.7/77.2 & 97.3/4.4/11.3/87.0 & 99.1/25.3/32.5/94.0 & 97.3/36.6/42.1/97.3 \\
regulator & 46.8/26.4/43.9 & 85.2/75.4/66.1 & 81.7/70.0/59.4 & 93.7/0.8/3.4/71.1 & 99.1/26.1/32.5/93.9 & 95.1/38.9/45.6/95.1 \\
rolled\_strip\_base & 97.3/98.6/94.4 & 99.4/99.7/97.6 & 99.1/99.5/97.1 & 98.9/10.8/17.4/95.2 & 99.7/37.5/43.7/98.8 & 99.0/51.9/55.9/99.0 \\
sim\_card\_set & 91.9/90.3/87.7 & 96.2/96.7/90.2 & 96.0/95.9/91.3 & 96.7/13.0/20.9/79.4 & 98.5/51.8/50.6/87.0 & 91.9/40.3/47.6/91.9 \\
switch & 89.3/91.3/82.1 & 93.1/94.8/86.0 & 97.7/98.1/92.3 & 99.4/55.3/58.8/92.6 & 99.5/57.0/59.0/95.5 & 96.9/54.9/56.5/96.9 \\
tape & 95.1/93.2/84.2 & 98.1/97.4/91.8 & 96.7/93.5/89.8 & 99.5/27.8/36.3/96.6 & 99.7/36.4/42.8/98.4 & 98.6/49.4/51.5/98.6 \\
terminalblock & 84.4/85.8/78.4 & 97.3/97.6/92.2 & 94.7/94.0/90.5 & 98.9/13.9/25.6/92.1 & 99.8/35.1/39.3/98.4 & 98.9/49.4/51.6/98.9 \\
toothbrush & 84.9/85.4/81.3 & 85.5/84.2/81.6 & 86.6/87.8/80.8 & 96.8/20.8/30.4/87.5 & 97.3/24.0/32.8/89.8 & 90.0/30.6/38.8/90.0 \\
toy & 79.7/82.3/80.6 & 86.5/90.2/82.6 & 88.3/91.2/83.6 & 96.4/7.0/13.4/77.2 & 97.3/17.0/25.8/89.4 & 94.1/26.7/33.3/94.1 \\
toy\_brick & 80.0/73.9/68.6 & 77.9/73.6/67.0 & 81.2/77.0/70.3 & 97.9/17.4/28.2/85.4 & 97.6/23.6/30.9/83.8 & 84.4/34.3/41.2/84.4 \\
transistor1 & 95.8/96.6/91.1 & 97.9/98.4/93.6 & 97.3/98.1/92.9 & 98.8/26.2/33.2/93.2 & 99.5/41.1/41.7/97.2 & 97.2/47.4/49.9/97.2 \\
u\_block & 85.4/76.7/69.7 & 93.1/90.2/81.3 & 94.9/92.3/84.9 & 99.0/19.5/26.2/91.5 & 99.6/32.6/40.9/95.7 & 98.2/50.1/54.3/98.2 \\
usb & 84.5/82.9/75.4 & 92.8/92.1/83.9 & 93.6/93.1/85.5 & 98.5/19.5/29.1/88.1 & 99.6/41.1/46.8/97.1 & 97.9/41.8/47.5/97.9 \\
usb\_adaptor & 78.3/70.3/67.2 & 83.8/78.7/70.8 & 83.8/77.4/71.2 & 97.0/5.8/12.1/81.9 & 97.3/19.2/26.5/81.6 & 90.3/28.6/36.4/90.3 \\
vcpill & 83.7/81.9/70.7 & 90.8/90.1/80.4 & 94.6/94.2/86.4 & 99.1/49.1/51.2/91.0 & 99.0/51.2/54.5/89.8 & 95.8/70.9/68.8/95.8 \\
wooden\_beads & 82.8/81.5/71.4 & 89.5/88.9/79.3 & 92.1/91.5/83.2 & 97.5/21.2/28.9/83.9 & 98.6/32.2/38.9/89.9 & 95.0/51.5/53.1/95.0 \\
woodstick & 79.7/70.4/61.8 & 85.7/77.9/70.0 & 86.9/77.9/70.0 & 96.6/39.5/45.6/81.3 & 98.5/42.8/48.5/89.5 & 92.6/47.8/52.8/92.6 \\
zipper & 97.5/98.4/94.2 & 99.4/99.6/97.1 & 98.6/99.3/96.1 & 97.5/21.0/26.1/92.0 & 99.2/56.1/59.8/97.2 & 96.8/61.8/62.5/96.8 \\
\midrule
Average & 81.6/77.3/73.4 & 90.2/88.2/81.0 & 90.6/88.3/81.7 & 97.6/17.9/25.1/85.9 & 98.9/34.6/39.9/92.3 & 99.0/42.5/47.3/94.8 \\
\bottomrule
\end{tabular}}
\caption{Image-level and pixel-level separate-class results on Real-IAD datasets.}
\label{tab:ourtab}
\end{table*}

\begin{table}[h]
\centering
\renewcommand{\arraystretch}{1.0}
\setlength\tabcolsep{5.0pt}
\resizebox{\linewidth}{!}{
\small 
\begin{tabular}{ccccc}
\toprule
Method&  Memory Usage&  Training Time&Testing Time \\
\midrule
Baseline&14174MiB &5.53h&9.3h\\
Ours&14340MiB&5.92h&9.9h\\
\bottomrule
\end{tabular}}
\caption{Comparison of Memory Usage and Computation Speed.}
\end{table}

\subsection{Implementation Details}
All input images are resized to 392 × 392 without additional data augmentation. We choose ViT-Base/14 (patchsize=14) pre-trained by DINOv2-R \cite{darcet2023vision} as the frozen part of encoder. The model is trained for 50,000 iterations for multi-class settings on the Real-IAD datasets on a single NVIDIA RTX4090 24GB GPU. The StableAdamW optimizer with AMSGrad is employed, using a learning rate of \(2 \times 10^{-3}\), \(\beta = (0.9, 0.999)\), and weight decay of \(1 \times 10^{-4}\). All hyperparameters in the experiment are set to 1 by default.

\subsection{Comparison with SOTA}
We compare our method with several state-of-the-art unsupervised anomaly detection approaches, including RD4AD \cite{deng2022anomaly}, UniAD \cite{you2022unified}, SimpleNet \cite{liu2023simplenet}, DeSTSeg \cite{zhang2023destseg}, DiAD \cite{he2024diffusion}, MambaAD \cite{hemambaad}, and Dinomaly \cite{guo2024dinomaly}, on the challenging Real-IAD dataset. The image-level and pixel-level results are presented in Table~\ref{tab:realiadsp} and Table~\ref{tab:realiadpx}, respectively.

As shown in Table~\ref{tab:realiadsp}, our method achieves image-level AUROC/AP/$F_1$-max scores of \textbf{91.0/88.6/82.1}, outperforming the previous best method (Dinomaly) by significant margins of \textbf{+1.7/+1.8/+1.9}, demonstrating the effectiveness of our proposed multi-class multi-view framework. Notably, our method surpasses these seven baselines across the vast majority of object categories, indicating superior generalization to diverse industrial objects.

In terms of pixel-level anomaly localization, as reported in Table~\ref{tab:realiadpx}, our method achieves \textbf{99.1/43.9/48.2/95.2} in AUROC/AP/$F_1$-max/AUPRO, outperforming the previous SOTA by \textbf{+0.3/+1.1/+1.1/+1.3}, respectively. This substantial improvement in AP and $F_1$-max shows that our model not only detects anomalies more accurately but also localizes them more precisely at the pixel level, especially in complex multi-class multi-view scenarios.

These results clearly validate the superiority of our approach in both image-level detection and pixel-level localization, achieving new SOTA for anomaly detection on Real-IAD.

\subsection{Visualization Results}
To further demonstrate the effectiveness of our method, we conduct the visualization experiments on Real-IAD datasets and show the anomaly map of all 30 categories in Figure \ref{vis}. The anomaly map are all randomly chosen. Each triplet of rows displays the input image, the pixel-level ground truth annotation, and the corresponding anomaly heatmap generated by our model. As shown, our method is capable of precisely localizing anomalies across diverse object types, even when the defects are small or subtle. This indicates that the model possesses strong generalization capability, fine-grained anomaly recognition ability, and effective suppression of normal patterns.

\subsection{Ablation Study}
Table \ref{tab:abiad} shows the ablation studies of three proposed components on the Real-IAD datasets. It can be obviously observed that adding all three components leads to the highest image-level performance, with AUROC, AP and $F_1$-max reaching 91.0\%, 88.6\% and 82.1\%, respectively. Pixel-level results follow a similar trend, with the best AUROC and AUPRO reaching 99.1\% and 95.2\% when all three components are used. However, AP and $F_1$-max for the pixel-level task are slightly lower in the configuration with all three components, compared to when only two components are used. Based on our analysis, this phenomenon may be caused by over-amplification effect of AAM. The AAM highlights anomalies by enhancing the abnormal regions and suppressing normal regions. However, at the pixel-level, the over-amplified anomaly signals may lead to overemphasis on certain local regions, causing the model to struggle with responding effectively to pixel-level anomalies, which might result in a decrease in AP and $F_1$-max.

\subsection{Computation Speed and Memory Usage}
In this section, we conduct a series of experiments to evaluate the impact of our proposed modules, semi-frozen encoder and anomaly amplification module, on computation speed and memory usage. All the experiments are conducted on a single NVIDIA RTX4090 24GB GPU. The table below shows the comparison of the baseline and our method on computation speed and memory usage when training and testing on Real-IAD datasets. The results indicate that adding Semi-Frozen Encoder and Anomaly Amplification Module does not significantly increase memory usage or noticeably reduce inference speed.

\subsection{Scaling of Encoder Architectures}
In this section, we explore how different scalings of frozen part of encoder influences the model's accuracy. Table \ref{tab:arch_realiad} presents a comparison of ViT architectures of different scales (Small, Base, and Large) on the Real-IAD industrial anomaly detection dataset, evaluating both image-level and pixel-level performance. As the model size increases, performance improves steadily. It is worth noting that our method already achieves state-of-the-art performance even when using ViT-Small as the backbone. These results also confirm the scalability and adaptability of our method across different Transformer backbone variants.

\begin{table}[h]
  \centering
  \large
  \resizebox{\linewidth}{!}{
    \begin{tabular}{cccccccc}
    \toprule
    \multirow{2}[2]{*}{Arch.} & \multicolumn{3}{c}{Image-level} & \multicolumn{4}{c}{Pixel-level} \\
    \cmidrule(r){2-4} \cmidrule(l){5-8} 
    & \multicolumn{1}{c}{AUROC} & \multicolumn{1}{c}{AP} & \multicolumn{1}{c}{$F_1$-max} & \multicolumn{1}{c}{AUROC} & \multicolumn{1}{c}{AP} & \multicolumn{1}{c}{$F_1$-max} & \multicolumn{1}{c}{AUPRO} \\
    \hline
    ViT-Small & 90.2  &  88.0 & 80.0  &  98.6 &  42.7 & 47.1  & 94.2 \\
    ViT-Base & 91.0 & 88.6 & 82.1 & 99.1 & 43.9 & 48.2 & 95.2 \\
    ViT-Large & 91.1 & 88.8 & 82.2 & 99.1 & 44.2 & 48.3 & 95.3 \\
    \bottomrule
    \end{tabular}
  }
  \caption{Scaling of ViT architectures on Real-IAD (\%).}
\label{tab:arch_realiad}
\end{table}

\subsection{Performance on Separate Training Settings}
To explore the performance of our model on separate training settings, we conduct the experiments where each category is associated with a unique model, whereas in the multiclass setting, all categories are trained by a single model. We select two highly representative methods, UniAD \cite{you2022unified} and MVAD \cite{he2024learning}, as baselines for comparison. The model is trained for 5000 iterations for each category. The results are shown in Table \ref{tab:ourtab}. It can be obviously observed that although our model is not specifically designed for separate-class training, it still achieves strong performance under this setting. Our method slightly outperforms the SOTA in image-level anomaly detection, and significantly surpasses it in pixel-level anomaly localization, particularly in terms of AP and $F_1$-max metrics.

\section{Conclusion}
In this paper, we introduce an encoder-decoder framework to train an unified model to solve multi-view anomaly detection. First, we propose a semi-frozen encoder to enable the model quickly adapt to the data distribution of different views. Moreover, we design an anomaly amplification module, which suppresses dominant normal patterns while amplifying anomalies. Finally, we introduce a cross-feature loss to locate anomalies in shallow layers and enhance them in deeper
layers, improving the model’s capacity for accurate and robust anomaly detection. Extensive experiments on Real-IAD indicate the superiority of our method on both image-level detection and pixel-level localization. 

\textbf{Future Work.} First, although the introduced components do not significantly increase the computational cost, training time and inference time, the transformer-based encoder-decoder framework remains relatively slower compared to other methods. Second, anomaly amplification module may over-amplify anomaly signals, causing the model to struggle with responding effectively to pixel-level anomalies, which results in a slight decrease in pixel-level AP and $F_1$-max metrics. We explore reducing the weight of the AAM module in the training phase. While this leads to improvements in the two evaluation metrics mentioned before, it ultimately causes a drop in overall performance. In future work, we will focus on improving the training and inference efficiency of the framework, as well as developing adaptive anomaly amplification strategies to balance pixel-level accuracy and overall performance.


\begin{thebibliography}{10}

\bibitem{bergmann2019mvtec}
Paul Bergmann, Michael Fauser, David Sattlegger, and Carsten Steger.
\newblock Mvtec ad--a comprehensive real-world dataset for unsupervised anomaly detection.
\newblock In {\em Proceedings of the IEEE/CVF conference on computer vision and pattern recognition}, pages 9592--9600, 2019.

\bibitem{bergmann2020uninformed}
Paul Bergmann, Michael Fauser, David Sattlegger, and Carsten Steger.
\newblock Uninformed students: Student-teacher anomaly detection with discriminative latent embeddings.
\newblock In {\em Proceedings of the IEEE/CVF conference on computer vision and pattern recognition}, pages 4183--4192, 2020.

\bibitem{chen2017multi}
Xiaozhi Chen, Huimin Ma, Ji~Wan, Bo~Li, and Tian Xia.
\newblock Multi-view 3d object detection network for autonomous driving.
\newblock In {\em Proceedings of the IEEE conference on Computer Vision and Pattern Recognition}, pages 1907--1915, 2017.

\bibitem{darcet2023vision}
Timoth{\'e}e Darcet, Maxime Oquab, Julien Mairal, and Piotr Bojanowski.
\newblock Vision transformers need registers.
\newblock {\em arXiv preprint arXiv:2309.16588}, 2023.

\bibitem{deng2022anomaly}
Hanqiu Deng and Xingyu Li.
\newblock Anomaly detection via reverse distillation from one-class embedding.
\newblock In {\em Proceedings of the IEEE/CVF Conference on Computer Vision and Pattern Recognition}, pages 9737--9746, 2022.

\bibitem{deng2024prioritized}
Huilin Deng, Hongchen Luo, Wei Zhai, Yang Cao, and Yu~Kang.
\newblock Prioritized local matching network for cross-category few-shot anomaly detection.
\newblock {\em IEEE Transactions on Artificial Intelligence}, 2024.

\bibitem{dong2024nng}
Hao Dong, Ga{\"e}tan Frusque, Yue Zhao, Eleni Chatzi, and Olga Fink.
\newblock Nng-mix: Improving semi-supervised anomaly detection with pseudo-anomaly generation.
\newblock {\em IEEE Transactions on Neural Networks and Learning Systems}, 2024.

\bibitem{gong2019memorizing}
Dong Gong, Lingqiao Liu, Vuong Le, Budhaditya Saha, Moussa~Reda Mansour, Svetha Venkatesh, and Anton van~den Hengel.
\newblock Memorizing normality to detect anomaly: Memory-augmented deep autoencoder for unsupervised anomaly detection.
\newblock In {\em Proceedings of the IEEE/CVF international conference on computer vision}, pages 1705--1714, 2019.

\bibitem{guo2023recontrast}
Jia Guo, Shuai Lu, Lize Jia, Weihang Zhang, and Huiqi Li.
\newblock Recontrast: Domain-specific anomaly detection via contrastive reconstruction.
\newblock {\em Advances in Neural Information Processing Systems}, 36:10721--10740, 2023.

\bibitem{guo2024dinomaly}
Jia Guo, Shuai Lu, Weihang Zhang, Fang Chen, Hongen Liao, and Huiqi Li.
\newblock Dinomaly: The less is more philosophy in multi-class unsupervised anomaly detection.
\newblock {\em arXiv preprint arXiv:2405.14325}, 2024.

\bibitem{hemambaad}
Haoyang He, Yuhu Bai, Jiangning Zhang, Qingdong He, Hongxu Chen, Zhenye Gan, Chengjie Wang, Xiangtai Li, Guanzhong Tian, and Lei Xie.
\newblock Mambaad: Exploring state space models for multi-class unsupervised anomaly detection.
\newblock In {\em The Thirty-eighth Annual Conference on Neural Information Processing Systems}, 2024.

\bibitem{he2024diffusion}
Haoyang He, Jiangning Zhang, Hongxu Chen, Xuhai Chen, Zhishan Li, Xu~Chen, Yabiao Wang, Chengjie Wang, and Lei Xie.
\newblock A diffusion-based framework for multi-class anomaly detection.
\newblock In {\em Proceedings of the AAAI conference on artificial intelligence}, volume~38, pages 8472--8480, 2024.

\bibitem{he2024learning}
Haoyang He, Jiangning Zhang, Guanzhong Tian, Chengjie Wang, and Lei Xie.
\newblock Learning multi-view anomaly detection.
\newblock {\em arXiv preprint arXiv:2407.11935}, 2024.

\bibitem{li2021cutpaste}
Chun-Liang Li, Kihyuk Sohn, Jinsung Yoon, and Tomas Pfister.
\newblock Cutpaste: Self-supervised learning for anomaly detection and localization.
\newblock In {\em Proceedings of the IEEE/CVF conference on computer vision and pattern recognition}, pages 9664--9674, 2021.

\bibitem{li2021center}
Daoming Li, Qinghua Tao, Jiahao Liu, and Huangang Wang.
\newblock Center-aware adversarial autoencoder for anomaly detection.
\newblock {\em IEEE Transactions on Neural Networks and Learning Systems}, 33(6):2480--2493, 2021.

\bibitem{liu2021anomaly}
Yixin Liu, Zhao Li, Shirui Pan, Chen Gong, Chuan Zhou, and George Karypis.
\newblock Anomaly detection on attributed networks via contrastive self-supervised learning.
\newblock {\em IEEE transactions on neural networks and learning systems}, 33(6):2378--2392, 2021.

\bibitem{liu2023simplenet}
Zhikang Liu, Yiming Zhou, Yuansheng Xu, and Zilei Wang.
\newblock Simplenet: A simple network for image anomaly detection and localization.
\newblock In {\em Proceedings of the IEEE/CVF conference on computer vision and pattern recognition}, pages 20402--20411, 2023.

\bibitem{pang2022zoom}
Youwei Pang, Xiaoqi Zhao, Tian-Zhu Xiang, Lihe Zhang, and Huchuan Lu.
\newblock Zoom in and out: A mixed-scale triplet network for camouflaged object detection.
\newblock In {\em Proceedings of the IEEE/CVF Conference on computer vision and pattern recognition}, pages 2160--2170, 2022.

\bibitem{roth2022towards}
Karsten Roth, Latha Pemula, Joaquin Zepeda, Bernhard Sch{\"o}lkopf, Thomas Brox, and Peter Gehler.
\newblock Towards total recall in industrial anomaly detection.
\newblock In {\em Proceedings of the IEEE/CVF conference on computer vision and pattern recognition}, pages 14318--14328, 2022.

\bibitem{su2015multi}
Hang Su, Subhransu Maji, Evangelos Kalogerakis, and Erik Learned-Miller.
\newblock Multi-view convolutional neural networks for 3d shape recognition.
\newblock In {\em Proceedings of the IEEE international conference on computer vision}, pages 945--953, 2015.

\bibitem{wang2024real}
Chengjie Wang, Wenbing Zhu, Bin-Bin Gao, Zhenye Gan, Jiangning Zhang, Zhihao Gu, Shuguang Qian, Mingang Chen, and Lizhuang Ma.
\newblock Real-iad: A real-world multi-view dataset for benchmarking versatile industrial anomaly detection.
\newblock In {\em Proceedings of the IEEE/CVF Conference on Computer Vision and Pattern Recognition}, pages 22883--22892, 2024.

\bibitem{wang2022mvster}
Xiaofeng Wang, Zheng Zhu, Guan Huang, Fangbo Qin, Yun Ye, Yijia He, Xu~Chi, and Xingang Wang.
\newblock Mvster: Epipolar transformer for efficient multi-view stereo.
\newblock In {\em European Conference on Computer Vision}, pages 573--591. Springer, 2022.

\bibitem{yang2023aide}
Dingkang Yang, Shuai Huang, Zhi Xu, Zhenpeng Li, Shunli Wang, Mingcheng Li, Yuzheng Wang, Yang Liu, Kun Yang, Zhaoyu Chen, et~al.
\newblock Aide: A vision-driven multi-view, multi-modal, multi-tasking dataset for assistive driving perception.
\newblock In {\em Proceedings of the IEEE/CVF International Conference on Computer Vision}, pages 20459--20470, 2023.

\bibitem{you2022unified}
Zhiyuan You, Lei Cui, Yujun Shen, Kai Yang, Xin Lu, Yu~Zheng, and Xinyi Le.
\newblock A unified model for multi-class anomaly detection.
\newblock {\em Advances in Neural Information Processing Systems}, 35:4571--4584, 2022.

\bibitem{yu2024tf}
Qianzi Yu, Kai Zhu, Yang Cao, Feijie Xia, and Yu~Kang.
\newblock Tf 2: Few-shot text-free training-free defect image generation for industrial anomaly inspection.
\newblock {\em IEEE Transactions on Circuits and Systems for Video Technology}, 2024.

\bibitem{zavrtanik2021draem}
Vitjan Zavrtanik, Matej Kristan, and Danijel Sko{\v{c}}aj.
\newblock Draem-a discriminatively trained reconstruction embedding for surface anomaly detection.
\newblock In {\em Proceedings of the IEEE/CVF international conference on computer vision}, pages 8330--8339, 2021.

\bibitem{zavrtanik2021reconstruction}
Vitjan Zavrtanik, Matej Kristan, and Danijel Sko{\v{c}}aj.
\newblock Reconstruction by inpainting for visual anomaly detection.
\newblock {\em Pattern Recognition}, 112:107706, 2021.

\bibitem{zhang2023destseg}
Xuan Zhang, Shiyu Li, Xi~Li, Ping Huang, Jiulong Shan, and Ting Chen.
\newblock Destseg: Segmentation guided denoising student-teacher for anomaly detection.
\newblock In {\em Proceedings of the IEEE/CVF Conference on Computer Vision and Pattern Recognition}, pages 3914--3923, 2023.

\end{thebibliography}

\end{document}